\documentclass[runningheads]{llncs}
\usepackage{graphicx}
\usepackage{hyperref}
\usepackage{tabularray}
\SetTblrInner{rowsep=0pt}
\usepackage{amsmath}
\usepackage{booktabs} 
\usepackage{xcolor}
\usepackage{tabularray}
\usepackage{amsfonts} 

\begin{document}
%

\title{MRScore: Evaluating Radiology Report Generation with LLM-based Reward System}

%
%
\author{Yunyi Liu\inst{1}\and
Zhanyu Wang\inst{1}\and
Yingshu Li\inst{1}\and
Xinyu Liang\inst{2}\and
Lingqiao Liu\inst{3}\and
Lei Wang\inst{4}\and
Luping Zhou\inst{1}
}
\authorrunning{Y. Liu. et al.}


\institute{The University of Sydney, Sydney, NSW, Australia\\
\email{\{yunyi.liu1,zhanyu.wang,yingshu li, luping.zhou\}@sydney.edu.au}\\
 \and
Guangzhou University of Chinese Medicine, Guangzhou, China\\
\email{xinyu.liang31@gmail.com}\\
\and
The University of Adelaide, Adelaide, SA, Australia\\
\email{lingqiao.liu@adelaide.edu.au}\\
\and
The University of Wollongong, Wollongong, NSW, Australia\\
\email{leiw@uow.edu.au}}





%
\authorrunning{F. Author et al.}
%

%
\maketitle              
\begin{abstract}
In recent years, automated radiology report generation has experienced significant growth. This paper introduces MRScore, an automatic evaluation metric tailored for radiology report generation by leveraging Large Language Models (LLMs). Conventional NLG (natural language generation) metrics like BLEU are inadequate for accurately assessing the generated radiology reports, as systematically demonstrated by our observations within this paper. To address this challenge, we collaborated with radiologists to develop a framework that guides LLMs for radiology report evaluation, ensuring alignment with human analysis. Our framework includes two key components: i) utilizing GPT to generate large amounts of training data, i.e., reports with different qualities, and ii) pairing GPT-generated reports as accepted and rejected samples and training LLMs to produce MRScore as the model reward. Our experiments demonstrate MRScore's higher correlation with human judgments and superior performance in model selection compared to traditional metrics. 
Our code and datasets will be available on GitHub.
\keywords{Radiology Report Generation  \and Evaluation metrics \and Large Language Models \and Reward Model.}
\end{abstract}

\section{Introduction}
Automated assessment of text generation systems, such as those used for radiology report generation, typically involves the comparison of the generated reports against the reference reports to gauge semantic accuracy. However, widely utilized natural language generation (NLG) metrics, such as the BLEU~\cite{papineni2002bleu}, primarily quantify n-gram matches, often overlooking the critical aspects of lexical and structural diversity crucial for preserving meaning. Recent studies~\cite{li2023comprehensive} have highlighted several shortcomings in n-gram-based evaluation metrics. \underline{First}, these metrics may often misjudge paraphrasing due to rigid pattern matching. For example, traditional metrics like BLEU and METEOR\cite{reimers2019sentencebert} may favour one expression over another despite semantic equivalence, penalizing deviations from the reference structure. To address this approaches like BERTScore~\cite{zhang2020bertscore} utilized contextualized token embedding to detect paraphrasing more effectively. \underline{Second}, NLG metrics often fail to capture complex diagnostic information in reports adequately. As a result, they are increasingly paired with clinic-relevant scores, such as F1 scores of clinic entities labeled by CheXbert~\cite{smit2020chexbert} or RadGraph~\cite{jain2021radgraph}, for comprehensive evaluation.
\underline{Third} and more importantly,  despite the existing efforts in improving the evaluation of report generation, these evaluation metrics do not well align with human judgment~\cite{li2023comprehensive}. To bridge this gap, a recent work~\cite{yu2023evaluating} proposed the RadCliQ score, which linearly combines BLEU, BERTScore, CHEXBERT, and RadGraph F1 while regressing combination weights from human-marked error scores to better align with human evaluation. However, RadCliQ's reliance on a limited set of expensive human-annotated training samples for learning poses a challenge, and its evaluation criteria tend to prioritize errors in clinical findings while neglecting linguistic aspects.

To propel advancements in this field, this study proposes MRScore, an innovative metric tailored specifically for evaluating automated radiology report generation. Developed in collaboration with professional radiologists, MRScore is grounded in a framework that elucidates the expert rules and priorities for report assessment. It harnesses the power of large language models (LLMs) to autonomously generate human-like evaluation samples, which are then used to train an LLM-based reward model for automated scoring. Compared to existing evaluation metrics, our approach offers several advantages.  Firstly, it significantly enhances alignment with human evaluations, ensuring a more accurate assessment of report quality. Secondly, it improves viability by facilitating the generation of large training sets without the need for human annotations\footnote{Our framework does not rely on human annotations for training; however, it requires one or two examples of human evaluation to be included in the prompt to guide LLM in generating training samples.}. Lastly, it enhances the model’s transparency and flexibility to evaluation rules, potentially facilitating adaptation to various report generation scenarios. To operationalize our approach, we first assessed
the capacity of GPT-4~\cite{OpenAI2023GPT4TR} in generating human-like evaluations based on our designed prompt and leveraged it to generate low-, middle-, and high-quality reports according to seven criteria for each of 1,000 ground truth reports, resulting in 3,000 scoring samples. These samples were paired as $<$accepted, rejected$>$ reports, with the quality of the `accepted’ report being higher than that of the `rejected’ report by a margin. These report pairs were then used to train a reward model built upon the pretrained Mistral-7B-instruct~\cite{jiang2023mistral} backbone, with the predicted reward serving as the final MRScore for report evaluation. To validate our model, we scored 100 sample reports generated by GPT-4V \cite{OpenAI2023GPT4TR} and compared these scores with those from existing evaluation methods. Our correlation calculations demonstrated that MRScore achieved higher alignment with human judgment than other metrics.

Our main contributions are summarized as follows.
\begin{itemize}
    \item[(1)] This paper identifies the capability of GPT-4 in generating human-like evaluations of radiology reports. Leveraging this insight, we propose a method to autonomously generate evaluation samples for training, enabling the rapid construction of large training sets at low costs. 
    \item[(2)] We introduce a framework for assessing automated radiology reports. This framework transforms the evaluation criteria used by radiologists into a comprehensive scoring system, considering seven key criteria. By incorporating human-like evaluation standards, our framework significantly improves the alignment of model outputs with human assessments.
    \item[(3)] More importantly, building upon our novel assessment framework, we propose MRScore, an LLM-based model designed for automated report scoring. Through extensive testing, MRScore outperforms other metrics in terms of correlation with human evaluations. This underscores the effectiveness of our approach in accurately assessing the quality of radiology reports.
\end{itemize}

\section{Method}\label{sec:method}
\begin{figure}[h!]
\centering
\includegraphics[width=0.8\textwidth]{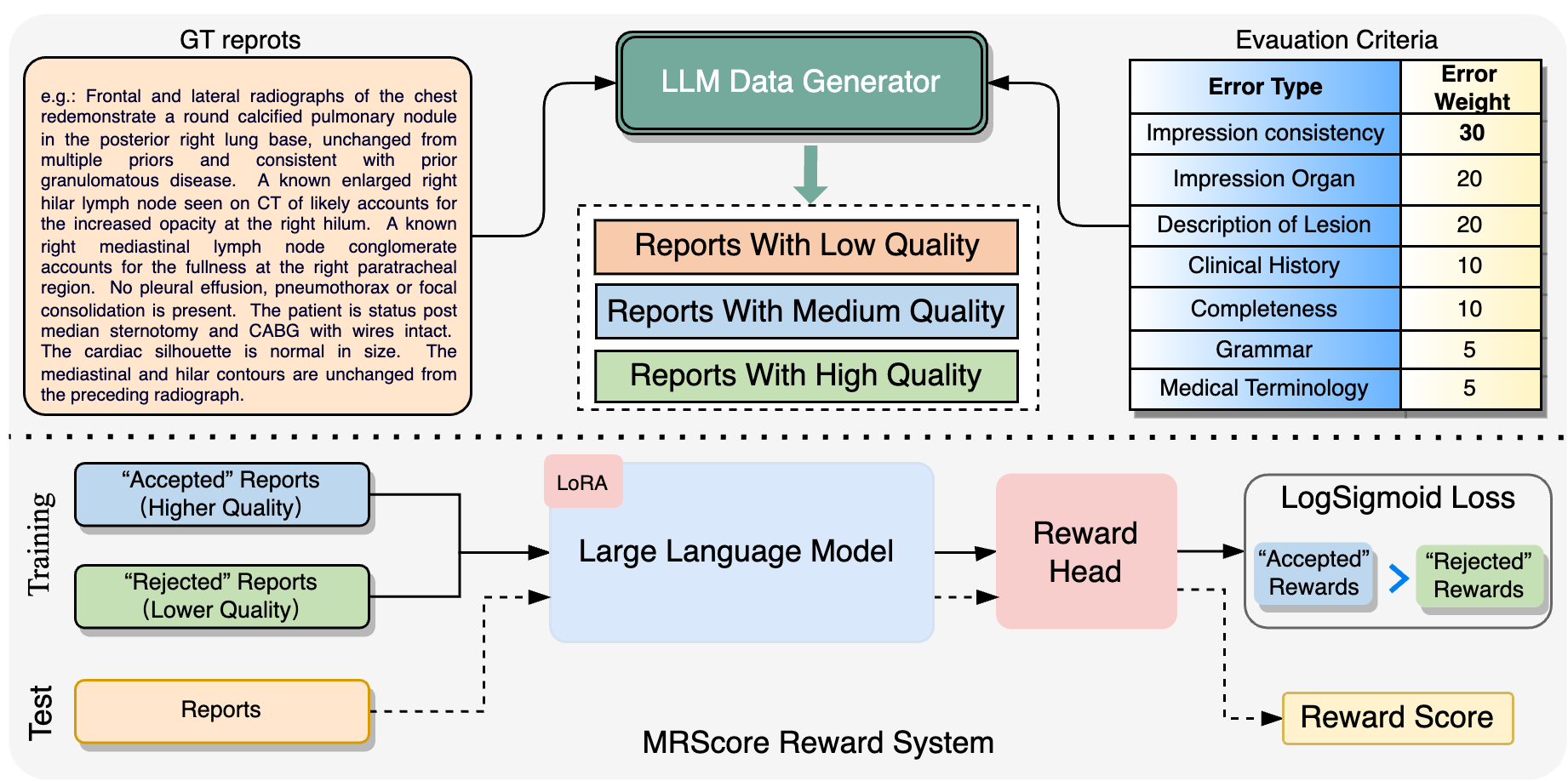}
\caption{Overview of MRScore. The upper portion illustrates the data generation process, while the lower portion represents the training process for the reward model using LoRA. In the lower portion, the solid line indicates the training phase while the dashed line indicates the inference phase.}
\label{fig:MRScoreModel}
\end{figure}

In the domain of automated radiology report generation, the efficacy of generated reports is assessed through a metric function $f(x, \hat{x})$, where $x$ represents the generated report and $\hat{x}$ is the reference report. Traditional evaluation metrics such as BLEU~\cite{papineni2002bleu}, ROUGE~\cite{lin2004rouge}, METEOR~\cite{banerjee2005meteor}, and CIDEr~\cite{vedantam2015cider} are commonly employed for this purpose. However, these metrics predominantly rely on n-gram overlap, which may not adequately capture semantic equivalence between the generated and reference texts. Therefore, we study to address the need for a more nuanced metric capable of accurately assessing the semantic content and clinical relevance of radiology reports.
The overview of our framework is presented in Figure~\ref{fig:MRScoreModel}, illustrating the design of the scoring framework, the acquisition of scoring data and training pairs, and the training of our model, which are elaborated in the following sections.


\subsection{Scoring Criteria Design and Scoring Dataset Generation}\label{subsec:ScoringCriteria}

\noindent \textbf{Report Evaluation Criteria}~~We collaborated closely with radiologists to develop a reliable scoring system based on seven fundamental items from their expertise and literature review, outlined in Table~\ref{table:errors1}. These items comprehensively cover both clinical findings and linguistic concerns, encompassing ``impression consistency", ``impression organs", ``description of lesions," ``clinical history", ``completeness", ``grammar", and ``medical terminology". A detailed explanation of each item is provided in Table~\ref{table:errors1}.
Our scoring system is error-based. Each item corresponds to an error type and is assigned a varied weight, as shown in Table~\ref{table:errors1}. The total score is calculated as follows, within the range of $[0, 100]$:
\begin{equation}
\text{Total\_score} = 100 - \sum_{i=1}^{7} S_i \times W_i.
\end{equation}
Here \(S_i\) represents the individual error score for the \(i\)-th error item, with a value of 1 if an error occurs and 0 if not. \(W_i\) denotes the weight for  the $i$-th error type, with specific values given in Table \ref{table:errors1}. 
\begin{table}[h!]
\centering
\caption{Table for error types and design detail.}
\label{table:errors1}
{\fontsize{5}{7}\selectfont
\begin{tabular}{|c|p{10cm}|} 
\hline
Error Type (Weight) & Explanation \\ \hline
Impression consistency (30) & Verify the presence of an `impression' section in the report first; its absence deems the report of inferior quality, as it contains crucial diagnostic details from the X-ray. A clear and accurate 'impression' is vital for radiologists to provide exceptional care. Crafting it demands substantial effort, as it involves more than just summarizing findings, with inaccuracies potentially diminishing the report's quality\cite{hartung2020create}. \\ \hline
Impression Organ (20) & Evaluate the precision of the impression section where present. A high-quality report should detail how surrounding organs are affected, highlighting the need for precise organ descriptions, a practice underscored by literature that recommends using subheadings for each organ and system in the findings\cite{ganeshan2018structured}. \\ \hline
Description of Lesion (20) & Review the report to ensure the anatomical organ related to the lesion is correctly identified. Check that it correctly describes the lesion's location, size, and opacity, as well as cardiovascular size and bone integrity, as per the ground truth. Descriptions of abnormal X-rays should include the lesion's precise anatomical location, distribution pattern, edges, shape, internal structure, and size, along with the quantity of lesions, their density, and the affected surroundings. Use standard terms for the lesion's location, describe its boundaries clearly, and measure the size of all significant lesions. For multiple lesions, record the size of the largest one, and describe the quantity and density accurately. Note any additional abnormalities within the scan area, and suggest further examination if the findings are inconclusive. \\ \hline
Clinical History (10) & Ensure that the predicted report accurately mentions any operation history, treatment, and family history as outlined in the ground truth. Providing clinical history is essential for justifying radiological exams and allows radiologists to fine-tune their analysis to address clinical queries, ultimately leading to diagnoses that integrate patient history and imaging findings\cite{clinicalhistory2017}. \\ \hline
Completeness (10) & Verify the completeness of the predicted report. Completeness is as essential as accuracy in radiology reports, reflecting the radiologist's expertise in providing detailed yet concise information vital for comprehensive patient care. \\ \hline
Grammar (5) & Ensure the predicted report is free from spelling errors and is clearly articulated. Proper grammar, accurate spelling, precise language, and coherent structure are pivotal in radiology reports for ensuring clarity\cite{wilcox2006written,pahadia2020radiology}. Since these reports significantly influence professional evaluations and patient care, meticulous proofreading is crucial to prevent any misinterpretation\cite{Radiology2000}. \\ \hline
Medical Terminology (5) & Verify the correct usage of medical terminology in the predicted report. This standardized language is crucial for clear communication among healthcare professionals, particularly for interpreting imaging reports and facilitating accurate diagnoses, treatment planning, and collaboration\cite{lukaszewicz2016art}. \\ \hline
\end{tabular}
}
\end{table}

\noindent \textbf{GPT-4' Evaluation}~~We utilized our established scoring system to craft prompts that encapsulate our evaluation criteria, guiding GPT-4 to assess radiology reports in a manner akin to humans. To validate the consistency between GPT-4 and human evaluations, both a trained radiologist and GPT-4 scored 100 model-generated reports of various qualities against their corresponding ground-truth reports randomly sampled from the MIMIC-CXR dataset. The Kendall's Tau correlations measuring the ordinal association between two quantities and the corresponding p-values of statistic tests are presented in Table~\ref{kendall_correlation}. Notably, GPT-4's evaluations exhibit a robust correlation with the radiologist's ratings, demonstrating a correlation coefficient of 0.531. The exceptionally small p-value of 5.98e-11 further confirms the statistical significance of this correlation. GPT-4's human correlation evaluation significantly outperforms common NLG metrics and clinically related scores.
\begin{table}
\centering
\caption{Verification of GPT-4's Capacity in Generating Human-like Ratings}
\label{kendall_correlation}
\begin{tabular}{l c}
\hline
Metrics & Kendall's Tau $\uparrow$ (P value $\downarrow$) \\
\hline
GPT-4 & 0.531 (5.98e-11) \\
BERTScore & 0.092 (2.04e-01) \\
RadGraph F1 & 0.06 (4.32e-01) \\
BLEU-4 & 0.048 (5.07e-01) \\
ROUGE & 0.120 (9.92e-02) \\
METEOR & 0.082 (2.58e-01) \\
CIDEr & 0.032 (6.54e-01) \\
\hline
\end{tabular}
\end{table}

\noindent \textbf{Scoring Dataset Generation}~~~After confirming GPT-4's proficiency in generating human-like evaluations of radiology reports, we proceeded to construct our scoring dataset. Using the GPT-4 API, we generated 3000 reports with varying qualities based on a random selection of 1000 ground-truth reports from the MIMIC-CXR datasets. Specifically, for each ground-truth report, GPT-4 generated three reports corresponding to three quality tiers (0-40, 40-70, and 70-100). This approach ensured a balanced score distribution, maintaining uniform coverage across the scoring spectrum. Following the removal of problematic reports, we obtained a refined dataset comprising 2994 reports, each accompanied by evaluation scores from GPT-4.

\subsection{LLM-based Reward Model}
Building upon our previous analysis, we present MRScore, an innovative evaluation metric honed within our novel evaluation framework. MRScore functions as an LLM-based reward model, trained by the technique of Reinforcement Learning with Human Feedback (RLHF) \cite{stiennon2020learning}. It fine-tunes a pretrained LLM model such as Mistral \cite{jiang2023mistral} and calibrates it to align with human evaluations via paired $<$accepted, rejected$>$ reports. Each pair represents two reports generated from the same ground-truth report, where the accepted report receives a higher GPT-4 score than the rejected one. Through training, our reward model learns to assign higher rewards to accepted reports. During inference, the predicted reward serves as the MRScore for evaluation. Below, we elaborate on our framework.


\noindent \underline{\textbf {Training Pairs Generation}}~~~ The training pairs were obtained using our GPT-4 scoring dataset built upon our designed error-based evaluation framework. Given the high correlation between GPT-4 scoring and human evaluations, this dataset effectively simulates human ranking of radiology reports. Each pair consists of an accepted sample $y_{w}$ and a rejected sample $y_{l}$ corresponding to the same ground-truth report $x$, where $y_{w}$ achieves a higher GPT-4 score than $y_{l}$.
Utilizing these pairs, an `accept' prompt $p(x, y_{w})$ and a `reject' prompt $p(x, y_{l})$ could be constructed to guide the model training,
where $p(\cdot)$ represents the prompt operation.  

\noindent \underline{\textbf{Backbone Model}}~~~We employed supervised fine-tuning on the Mistral-7B-instruct backbone \cite{jiang2023mistral}, utilizing Low-Rank Adaptation (LoRA) \cite{hu2022lora} for parameter-efficient fine-tuning (PEFT) \cite{houlsby2019parameter} to calibrate our reward model by human-like evaluations. We selected Mistral-7B-instruct as our base language model due to its strong language understanding capabilities and proficiency in comprehension tasks. Despite its relatively modest size of 7 billion parameters, Mistral-7B-instruct outperformed other models in our experiments, justifying its efficiency and power as a choice for our framework.


\noindent \underline{\textbf{Objective}}~~~The reward model is designed to predict human preferences accurately. Typically, this involves establishing a reward function the model seeks to optimize. In the case of MRScore, the reward function relies on the rankings of radiology reports, enabling the model to learn to discern and predict which report is preferred within each report pair. The objective function is given in Section~\ref{subsec:LossFunction}.

\noindent \underline{\textbf {Fine-Tuning}}~~~During fine-tuning, the pretrained LLM backbone model is tailored to suit the nuances of reward prediction, ensuring its outputs align with anticipated human evaluations. In our MRScore training, we fine-tuned the Mistral-7B-instruct model using our training pairs, instructing it to differentiate between higher- and lower-quality reports while respecting the score margin between them: $margin = score_{accpet} - score_{reject}$. A larger margin indicates a more pronounced quality discrepancy between the two reports, while a smaller margin suggests a lesser difference.
On top of the original Mistral model, we incorporate a reward head—a linear projection layer—that maps the LLM output features to a single dimension. Subsequently, a sigmoid activation function is applied to scale the output between 0 and 1, where 1 signifies the highest reward and 0 represents the lowest. This reward score is then utilized to evaluate the quality of the generated reports, serving as our MRScore. 
The whole process is illustrated in the lower portion of Fig.~\ref{fig:MRScoreModel}.


\subsection{Loss Function}\label{subsec:LossFunction}
The overall loss function for our reward model is depicted in Equation~\ref{equ:loss}. Here, $\gamma_{\theta}(\cdot)$ denotes the reward function to be learned, while $\theta$ represents the learnable parameters of our reward model. Recall that $x$ represents the ground truth report; $y_w$ signifies reports with higher scores; and $y_l$ denotes reports with lower scores. The variable $m$ represents the score margin to differentiate between higher and lower-scored reports. $\log (\cdot)$ refers to the logistic function. $D$ stands for the reference dataset, and  $K$ for the batch size utilized in training.
\begin{equation}
\label{equ:loss}
\operatorname{loss}(\theta) = -\frac{1}{\binom{K}{2}} \sum_{(x, y_w, y_l) \sim D} \left[ \log \left( \sigma\left( \gamma_{\theta}(x, y_w) - \gamma_{\theta}(x, y_l) - m\right) \right) \right]
\end{equation}
According to Equation~\ref{equ:loss}, if the reward for a high-quality report does not exceed that for a low-quality report by the specified margin, a penalty will be incurred. In this way, our reward model is calibrated to align with human evaluation via our scoring dataset.


\section{Experiments and Result}
\subsection{Evaluation Dataset }
We assessed the effectiveness of MRScore by examining its alignment with expert radiologist evaluations, ensuring its predictions are closely correlated with human expert rankings. To achieve this, we curated an evaluation dataset comprising 100 reports generated by GPT-4V \cite{OpenAI2023GPT4TR}, a robust multi-modal LLM renowned for its ability to process image inputs seamlessly. Recent evaluations have demonstrated GPT-4V's proficiency in understanding medical images and its capability to generate high-quality radiology reports \cite{li2023comprehensive}, making it our choice for the evaluation dataset due to its diverse generation capabilities.
These reports were generated by inputting 100 randomly selected CXR images from the MIMIC-CXR dataset directly into GPT-4V for generation.\footnote{Please note that these 100 reports were directly generated by GPT-4V using images, different from the 100 reports used in Section~\ref{subsec:ScoringCriteria} to verify the alignment of GPT-4 scores with human evaluations.} Subsequently, a radiologist rated these 100 generated reports against their ground truth reports from MIMIC-CXR by assigning one of three quality levels: high, mid, or low. These quality levels correspond to the scores of 90, 60, and 30, respectively.

\subsection{Experiment Result}
Table~\ref{tau} shows the human correlations of our MRScore against other NLG metrics including BLEU-4, METEOR, and CIDEr, and clinical scores like BERTScore and RadGraph F1. We also compared with RadCliQ-based metrics that were regressed from human-annotated error scores.
Our results underscore MRScore's robust alignment with human judgments, as demonstrated by Kendall's Tau (0.178) and Spearman's coefficient (0.221), surpassing traditional NLG metrics and clinical scores. While conventional NLG metrics display insignificant correlations, clinical scores such as BERTScore and RadGraph F1 exhibit statistically significant correlations with human evaluations, supported by lower p-values. This aligns with our comprehensive evaluation criteria, which extend beyond rigid pattern matching to encompass clinical findings, thus contributing to the observed correlations. Another noteworthy observation pertains to RadCliQ-based metrics, which integrate measurements from BLEU, BERTScore, ChestXbert, and RadGraph F1. It is intriguing to note that these metrics exhibit a significant correlation with human evaluations in our study, despite being derived from human-annotated error scores using a distinct scoring criterion. We attribute this correlation to the shared emphasis on clinical findings in both RadCliQ and our studies.
\begin{table}[h!]
\centering
\caption{Assessment of Human Correlations on Evaluation Dataset}
\resizebox{\linewidth}{!}{
\begin{tabular}{l|c|c|c|c|c|c|c|c|c}
\hline
                        & BLEU-4    & ROUGE      & METEOR    & CIDEr     & BERTScore    & RadGraph F1      & RadCliQ-V0 & RadCliQ-V1 & \textbf{MRScore}   \\ \hline
P Value $\downarrow$                 & 0.688     & 0.429         & 0.463     & 0.503     & 0.0446        & 0.071                    & 0.018         & 0.016       & \textbf{0.002}     \\  
Kendall’s Tau $\uparrow$           & 0.032     & 0.063         & 0.059     & 0.053     & 0.159         & 0.144                   & -0.189        & -0.193      & \textbf{0.250}     \\ \hline
P Value $\downarrow$                 & 0.677     & 0.484         & 0.460     & 0.422     & 0.045        & 0.080                    & 0.0175       & 0.0015      & \textbf{0.002}     \\  
Spearman $\uparrow$          & 0.042     & 0.071         & 0.074     & 0.081     & 0.200         & 0.176                   & -0.236        & -0.241      & \textbf{0.304}     \\ \hline
\end{tabular}
}
\label{tau}
\end{table}

\begin{figure}[h!]
\includegraphics[width=0.9\textwidth]{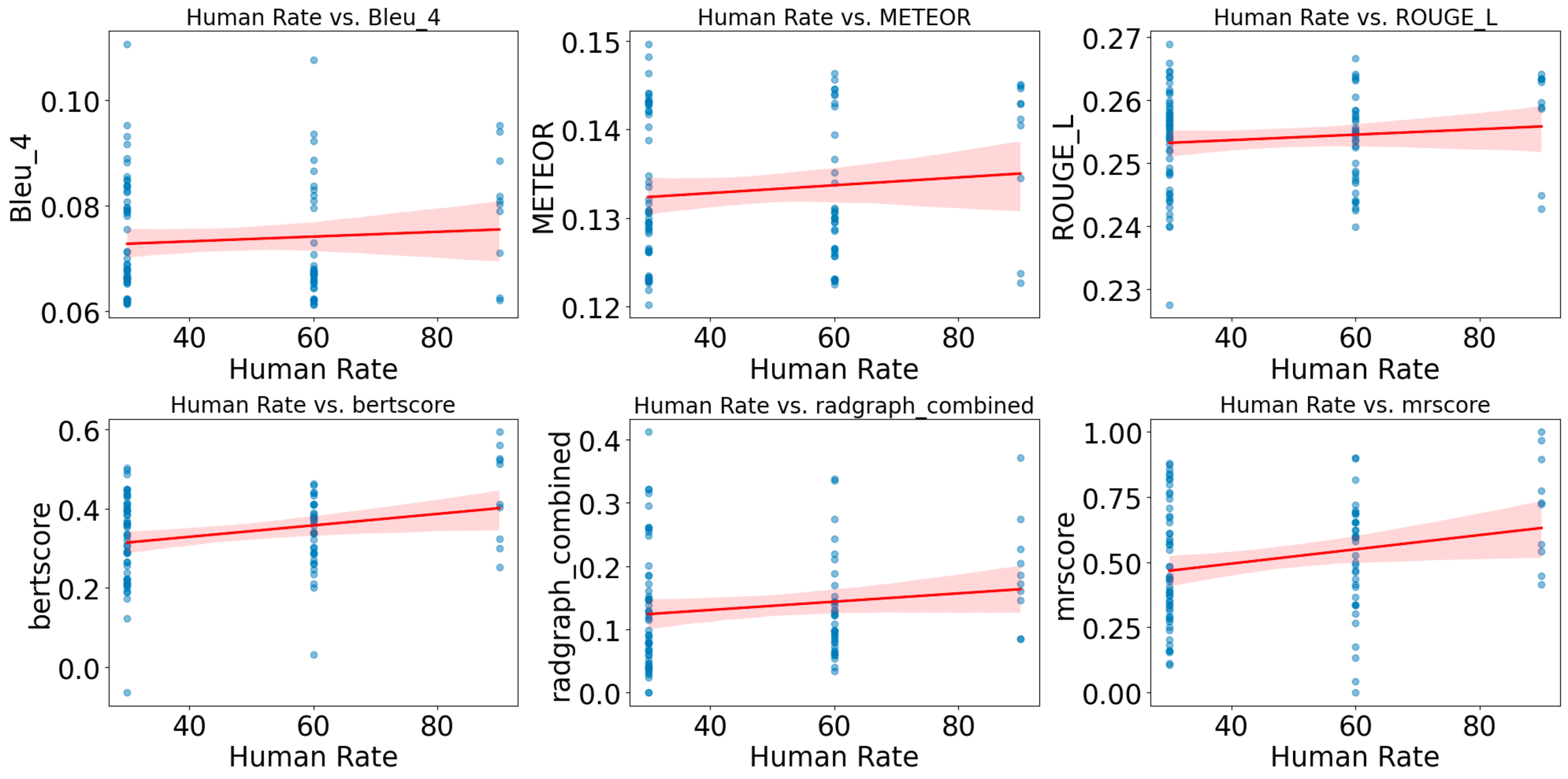}
\centering
\caption{Correlations between varied evaluation metrics and radiologist scores}
\label{fig:scatter_graph}
\end{figure}

The scatter plots in Fig.~\ref{fig:scatter_graph} offer insights into the comparisons between different scoring metrics and human evaluation ratings for radiology reports. Each plot corresponds to a specific metric, with data points illustrating individual report scores plotted against human ratings. The correlation trends are discernible through lines with shaded areas representing confidence intervals. Notably, MRScore demonstrates the most positive correlation with human ratings, suggesting a close alignment with professional evaluations of radiology reports. Traditional NLG metrics like BLEU-4, METEOR, and ROUGE show some positive correlation with human ratings but with a greater spread, indicating variability in their alignment. BERTScore also shows a positive correlation, albeit to a lesser extent than MRScore. Similarly, RadGraph F1 displays a positive trend but not as strong as MRScore. Overall, MRScore stands out as a promising metric for aligning with human judgment, potentially indicating its effectiveness.
%

Table~\ref{model_compare} shows the performance of our reward model using various LLMs, including Phi-2~\cite{li2023phi2}, Gemma-2b-it~\cite{gemma_2024}, and Gemma-7b-it~\cite{gemma_2024}, alongside our chosen Mistral-7b~\cite{jiang2023mistral} as the backbone models. Once again, both Kendall and Spearman correlation coefficients are utilized for assessment.
Among these models, Mistral-7b, with 6.8M trainable parameters, demonstrates the highest consistency with human ratings. It achieves a Kendall correlation of 0.179 and a Spearman correlation of 0.220, better than Gemma-7b-it~\cite{gemma_2024} that has similar amount of trainable parameters. As a result, we opted for Mistral as our backbone LLM.
\begin{table}
\centering
\caption{Human Correlations of MRScore Using Different LLM Backbones}
\label{model_compare}
\begin{tblr}{
  colsep=0.7mm,
  column{even} = {c},
  column{3} = {c},
  hline{1-5,6} = {-}{},
  vline{2-4} = {-}{}
}
Model       & Trainable params (\%) & Kendall $\uparrow$(P value $\downarrow$) & Spearman $\uparrow$(P value $\downarrow$) \\
Phi-2~\cite{li2023phi2}     & 5.2M (0.197)                     & 0.144 (0.069)          & 0.178 (0.075)       \\
Gemma-2b-it~\cite{gemma_2024} & 1.8M (0.073)                     & 0.135 (0.091)          & 0.169 (0.092)       \\
Gemma-7b-it~\cite{gemma_2024} & 6.4M (0.075)                     & 0.170 (0.034)          & 0.209 (0.037)       \\
Mistral-7b~\cite{jiang2023mistral}  & 6.8M (0.096)               & \textbf{0.250} (\textbf{0.002})          & \textbf{0.304} (\textbf{0.002})      
\end{tblr}
\end{table}

\section{Conclusion}
This paper presents a pioneering endeavor to train LLM models to evaluate radiology reports in a manner akin to human experts. Leveraging reinforcement learning with human feedback, we fine-tuned pretrained LLM models to achieve human-like scoring capabilities. To overcome the cost and time limitations associated with acquiring large-scale human evaluations for training, we harnessed the power of GPT-4 to generate human-like evaluation samples, according to a meticulously designed scoring system comprising seven critical evaluation criteria. Our findings demonstrate the efficacy of our approach, with our MRScore exhibiting a remarkable correlation with human evaluations, surpassing other traditional evaluation metrics. This achievement advances the capabilities of AI to offer a more accurate and cost-effective evaluation for automated radiology report generation.

\bibliographystyle{splncs04}
\newpage
\bibliography{refs}

\begin{thebibliography}{10}
\providecommand{\url}[1]{\texttt{#1}}
\providecommand{\urlprefix}{URL }
\providecommand{\doi}[1]{https://doi.org/#1}

\bibitem{banerjee2005meteor}
Banerjee, S., Lavie, A.: Meteor: An automatic metric for mt evaluation with improved correlation with human judgments. In: Proceedings of the acl workshop on intrinsic and extrinsic evaluation measures for machine translation and/or summarization. pp. 65--72 (2005)

\bibitem{ganeshan2018structured}
Ganeshan, D., Duong, P.A.T., Probyn, L., Lenchik, L., McArthur, T.A., Retrouvey, M., Ghobadi, E.H., Desouches, S.L., Pastel, D., Francis, I.R.: Structured reporting in radiology. Academic radiology  \textbf{25}(1),  66--73 (2018)

\bibitem{gemma_2024}
Gemma~Team, T.M., Hardin, C., Dadashi, R., Bhupatiraju, S., Sifre, L., Rivière, M., Kale, M.S., Love, J., Tafti, P., Hussenot, L., et~al.: Gemma  (2024). \doi{10.34740/KAGGLE/M/3301}, \url{https://www.kaggle.com/m/3301}

\bibitem{Radiology2000}
Hall, F.M.: The title of the article. \url{https://www.ajronline.org/doi/10.2214/ajr.175.5.1751239} (2000), accessed: 2023-02-24

\bibitem{hartung2020create}
Hartung, M.P., Bickle, I.C., Gaillard, F., Kanne, J.P.: How to create a great radiology report. RadioGraphics  \textbf{40}(6),  1658--1670 (2020)

\bibitem{houlsby2019parameter}
Houlsby, N., Giurgiu, A., Jastrzebski, S., Morrone, B., De~Laroussilhe, Q., Gesmundo, A., Attariyan, M., Gelly, S.: Parameter-efficient transfer learning for nlp. In: International conference on machine learning. pp. 2790--2799. PMLR (2019)

\bibitem{hu2022lora}
Hu, E.J., Shen, Y., Wallis, P., Allen-Zhu, Z., Li, Y., Wang, S., Wang, L., Chen, W.: Lo{RA}: Low-rank adaptation of large language models. In: International Conference on Learning Representations (2022), \url{https://openreview.net/forum?id=nZeVKeeFYf9}

\bibitem{jain2021radgraph}
Jain, S., Agrawal, A., Saporta, A., Truong, S.Q., Duong, D.N., Bui, T., Chambon, P., Zhang, Y., Lungren, M.P., Ng, A.Y., et~al.: Radgraph: Extracting clinical entities and relations from radiology reports. arXiv preprint arXiv:2106.14463  (2021)

\bibitem{jiang2023mistral}
Jiang, A.Q., Sablayrolles, A., Mensch, A., Bamford, C., Chaplot, D.S., et~al.: Mistral 7b (2023)

\bibitem{li2023comprehensive}
Li, Y., Liu, Y., Wang, Z., Liang, X., Liu, L., Wang, L., Cui, L., Tu, Z., Wang, L., Zhou, L.: A comprehensive study of gpt-4v's multimodal capabilities in medical imaging. medRxiv pp. 2023--11 (2023)

\bibitem{li2023phi2}
Li, Y., Bubeck, S., Eldan, R., Giorno, A.D., Gunasekar, S., Lee, Y.T.: Textbooks are all you need ii: phi-1.5 technical report (2023)

\bibitem{lin2004rouge}
Lin, C.Y.: Rouge: A package for automatic evaluation of summaries. In: Text summarization branches out. pp. 74--81 (2004)

\bibitem{lukaszewicz2016art}
Lukaszewicz, A., Uricchio, J., Gerasymchuk, G.: The art of the radiology report: practical and stylistic guidelines for perfecting the conveyance of imaging findings. Canadian Association of Radiologists Journal  \textbf{67}(4),  318--321 (2016)

\bibitem{OpenAI2023GPT4TR}
OpenAI: Gpt-4 technical report. ArXiv  \textbf{abs/2303.08774} (2023), \url{https://api.semanticscholar.org/CorpusID:257532815}

\bibitem{pahadia2020radiology}
Pahadia, M., Khurana, S., Geha, H., Deahl, S.T.I.: Radiology report writing skills: A linguistic and technical guide for early-career oral and maxillofacial radiologists. Imaging Science in Dentistry  \textbf{50}(3), ~269 (2020)

\bibitem{papineni2002bleu}
Papineni, K., Roukos, S., Ward, T., Zhu, W.J.: Bleu: a method for automatic evaluation of machine translation. In: Proceedings of the 40th annual meeting of the Association for Computational Linguistics. pp. 311--318 (2002)

\bibitem{clinicalhistory2017}
Radiology, C.C.: Why clinical history is essential for diagnoses. \url{https://radiologyblog.cincinnatichildrens.org/why-clinical-history-essential-for-diagnoses/} (2017), accessed: 2023-02-24

\bibitem{reimers2019sentencebert}
Reimers, N., Gurevych, I.: Sentence-bert: Sentence embeddings using siamese bert-networks. In: Inui, K., Jiang, J., Ng, V., Wan, X. (eds.) Proceedings of the 2019 Conference on Empirical Methods in Natural Language Processing and the 9th International Joint Conference on Natural Language Processing, {EMNLP-IJCNLP} 2019, Hong Kong, China, November 3-7, 2019. pp. 3980--3990. Association for Computational Linguistics (2019). \doi{10.18653/v1/D19-1410}, \url{https://doi.org/10.18653/v1/D19-1410}

\bibitem{smit2020chexbert}
Smit, A., Jain, S., Rajpurkar, P., Pareek, A., Ng, A.Y., Lungren, M.P.: Chexbert: combining automatic labelers and expert annotations for accurate radiology report labeling using bert. arXiv preprint arXiv:2004.09167  (2020)

\bibitem{stiennon2020learning}
Stiennon, N., Ouyang, L., Wu, J., Ziegler, D., Lowe, R., Voss, C., Radford, A., Amodei, D., Christiano, P.F.: Learning to summarize with human feedback. Advances in Neural Information Processing Systems  \textbf{33},  3008--3021 (2020)

\bibitem{vedantam2015cider}
Vedantam, R., Lawrence~Zitnick, C., Parikh, D.: Cider: Consensus-based image description evaluation. In: Proceedings of the IEEE conference on computer vision and pattern recognition. pp. 4566--4575 (2015)

\bibitem{wilcox2006written}
Wilcox, J.R.: The written radiology report. Applied Radiology  \textbf{35}(7) (2006)

\bibitem{yu2023evaluating}
Yu, F., Endo, M., Krishnan, R., Pan, I., Tsai, A., Reis, E.P., Fonseca, E.K.U.N., Lee, H.M.H., Abad, Z.S.H., Ng, A.Y., et~al.: Evaluating progress in automatic chest x-ray radiology report generation. Patterns  \textbf{4}(9) (2023)

\bibitem{zhang2020bertscore}
Zhang, T., Kishore, V., Wu, F., Weinberger, K.Q., Artzi, Y.: Bertscore: Evaluating text generation with {BERT}. In: 8th International Conference on Learning Representations, {ICLR} 2020, Addis Ababa, Ethiopia, April 26-30, 2020. OpenReview.net (2020), \url{https://openreview.net/forum?id=SkeHuCVFDr}

\end{thebibliography}
\end{document}